\documentclass[runningheads]{llncs}
\usepackage{graphicx}

\usepackage{tikz}
\usepackage{comment} 
\usepackage{amsmath,amssymb} 
\usepackage{color}
\usepackage{bm}
\usepackage{multirow}
\usepackage[colorlinks,citecolor=red,citecolor=green]{hyperref}

\begin{document}
\pagestyle{headings}
\mainmatter
\def\ECCVSubNumber{100}  

\title{Enhanced Quadratic Video Interpolation} 

\titlerunning{Enhanced Quadratic Video Interpolation}
%
\author{Yihao Liu\inst{1,2}\thanks{The first two authors are co-first authors.} \and
Liangbin Xie\inst{1,2}$^{\star}$ \and
Li Siyao\inst{3} \and \\
Wenxiu Sun\inst{3} \and
Yu Qiao\inst{1} \and
Chao Dong\inst{1}}
\authorrunning{Liu et al.}
%
\institute{ShenZhen Key Lab of Computer Vision and Pattern Recognition, SIAT-SenseTime Joint Lab, Shenzhen Institutes of Advanced Technology, Chinese Academy of Sciences\\ \and
	University of Chinese Academy of Sciences \and SenseTime Research \\
	\email{\{lb.xie, yu.qiao, chao.dong\}@siat.ac.cn}, \email{liuyihao14@mails.ucas.ac.cn} \\
	\email{\{lisiyao1,sunwenxiu\}@tetras.ai}}

\maketitle

\begin{abstract}
With the prosperity of digital video industry, video frame interpolation has arisen continuous attention in computer vision community and become a new upsurge in industry. Many learning-based methods have been proposed and achieved progressive results. Among them, a recent algorithm named quadratic video interpolation (QVI) achieves appealing performance. It exploits higher-order motion information (\textit{e.g.} acceleration) and successfully models the estimation of interpolated flow. However, its produced intermediate frames still contain some unsatisfactory ghosting, artifacts and inaccurate motion, especially when large and complex motion occurs. In this work, we further improve the performance of QVI from three facets and propose an enhanced quadratic video interpolation (EQVI) model. In particular, we adopt a rectified quadratic flow prediction (RQFP) formulation with least squares method to estimate the motion more accurately. Complementary with image pixel-level blending, we introduce a residual contextual synthesis network (RCSN) to employ contextual information in high-dimensional feature space, which could help the model handle more complicated scenes and motion patterns. Moreover, to further boost the performance, we devise a novel multi-scale fusion network (MS-Fusion) which can be regarded as a learnable augmentation process. The proposed EQVI model won the first place in the AIM2020 Video Temporal Super-Resolution Challenge. Codes are available at \url{https://github.com/lyh-18/EQVI}.

\end{abstract}

\begin{figure}[h]
	\centering
	\includegraphics[width=12cm]{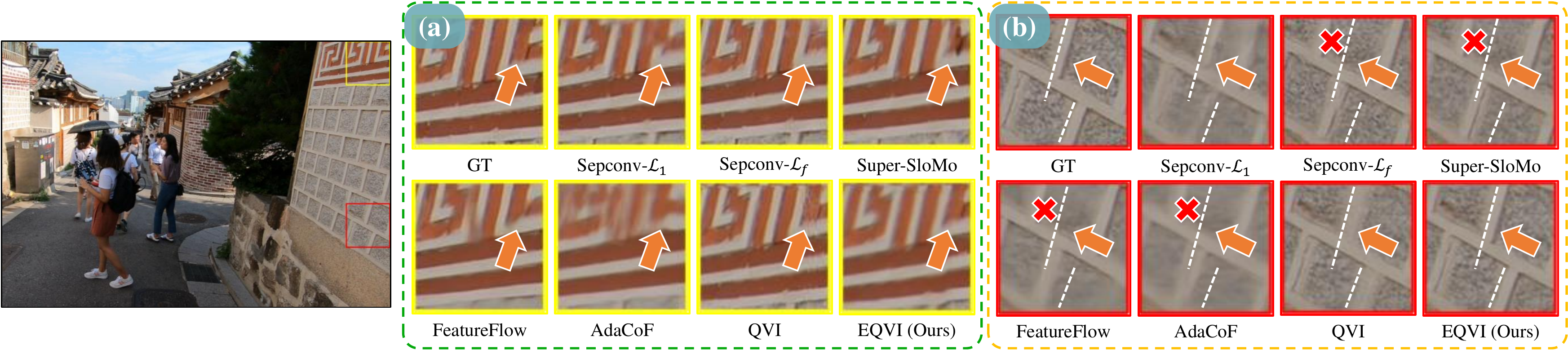}
	\caption{Video frame interpolation examples. Compared with other state-of-the-art methods, EQVI can generate more visually pleasing intermediate frames with more accurate textures and fewer artifacts. In patch (a), the results of other methods are prone to be blurry or contain artifacts in regions with complex structures (indicated by the arrows); In patch (b), other methods may result in blur and misalignment, due to the inaccurate motion estimation (indicated by the dashed lines and red crosses).
	}
	\label{fig:pageone}
\end{figure}

\section{Introduction}
Video frame interpolation has attracted increasing attention in recent years, due to its urgent needs in video enhancement market. It is intrinsically an ill posed problem that predicts intermediate/missing frames from a low frame-rate video. With the help of deep learning techniques, recent video interpolation algorithms have achieved significant improvement in both evaluation metrics and visual quality. To further promote the progress, the AIM 2020 held the challenge on video interpolation \cite{son2020aim_VTSR}. Our team won the challenge with the proposed enhanced quadratic video interpolation (EQVI) method. We will describe our winning entry in this paper, and compare with other state-of-the-arts.

Existing deep-learning-based methods can be roughly divided into two groups. 
The first one, which is summarized as “flow-based” method, tries to estimate the optical flow of the intermediate frame. Then, the corresponding pixels from adjacent frames are warped according to the estimated flow.
The representative methods are DVF\cite{voxelflow}, Super-SloMo\cite{superslomo}, DAIN\cite{dain} and QVI\cite{xu2019quadratic}. 
While the second group, concluded as “kernel-based” method, includes AdaConv\cite{adaconv}, Sepconv\cite{sepconv} and AdaCoF\cite{adacof}, which directly generate isolated sampling kernel maps for each pixel of the intermediate frame.
As most of the above methods perform video interpolation with only two adjacent frames, they inevitably adopt the linear motion assumption, which does not always hold in real scenarios. Among them, QVI model \cite{xu2019quadratic} applies a quadratic interpolation formula for predicting more accurate intermediate frames.

Our method is built upon QVI, and improves it from three aspects. 1) From the aspect of formulation, we refine the original quadratic flow prediction module by rectifying it with least squares method. This rectified quadratic flow prediction formula (RQFP) could correct the estimation error caused by calculating with incomplete data samples, and improve the accuracy of the interpolated optical flow when the frame motion satisfies the quadratic assumption. 2) To employ more information, we introduce another residual contextual synthesis network (RCSN), which can incorporate contextual information from pre-extracted high-dimensional features (ResNet-18 \cite{resnet}). This additional module could alleviate the problem of inaccurate motion estimation and object occlusion, and further push the performance by at least 0.18dB. 3) As a post-processing step, we propose a novel learnable augmentation module, named multi-scale fusion network. It is based on the observation that the performance can be further boosted by fusing results of different resolution levels. This component could improve PSNR value, but at the expense of SSIM. These three strategies are complementary to each other, thus can be adopted independently or simultaneously.

To maximize the final performance, we propose an efficient training strategy to sequentially use the above components. First, we train a standard QVI as our baseline. Then the RCSN is added and finetuned to exploit the contextual information in high-dimensional feature space. After that, we adopt the RQFP formula to rectify the interpolated optical flow. \footnote{RQFP has no trainable parameters and only rectifies the formula of intermediate flow estimation. However, it requires matrix multiplication and costs more training time, so we adopt it after RCSN is equipped to speed up the entire learning process.} At last, we employ multi-scale fusion network to further push the PSNR results.

Our full model, named Enhanced Quadratic Video Interpolation (EQVI), won the first place in AIM2020 VTSR challenge, and is superior to the second and third place by 0.08dB and 0.5dB, respectively. The proposed EQVI improves the original QVI by 0.38dB, and outperforms all recent state-of-the-art methods on REDS\_VTSR dataset (see Table \ref{Tab:comparison}). We have also conducted extensive experiments to evaluate each proposed component.

\section{Related Work}
Video frame interpolation is a long-standing topic and has been extensively studied in the computer vision community. Recently, a variety of learning-based algorithms have been developed for video frame interpolation.

In 2016, Niklaus \textit{et al.} \cite{adaconv} proposed a spatially-adaptive convolution to tackle video frame interpolation. They postulated that the per-pixel interpolation can be modeled with a convolutional operation over the corresponding patches from the two input frames. Subsequently, Sepconv \cite{sepconv} was introduced to separately estimate the convolutional kernels in different orientations. Afterwards, Liu \textit{et al.} \cite{voxelflow} constructed an encoder-decoder to predict the 3D voxel flow and linearly synthesize the desired intermediate frame. Combined with neural network, PhaseNet \cite{meyer2018phasenet} was designed to robustly handle challenging scenarios with larger motions. To handle multi-frame video interpolation, Jiang \textit{et al.} \cite{superslomo} proposed Super-SloMo, which jointly solves interpolation and occlusion estimation. For better blending the two warped frames, Niklaus \textit{et al.} \cite{niklaus2018context} presented a context-aware synthesis approach that warps not only the input frames but also the pixel-wise contextual information. Based on the observation that closer objects should be preferably synthesized, the depth information \cite{dain} was exploited to further facilitate interpolation.
The above works generally adopt the linear motion assumption, which is not a good approximation for real-world motions. To improve the accuracy of motion estimation, Xu \textit{et al.} proposed a quadratic video interpolation method to exploit both the velocity and acceleration information. In this paper, we improve QVI in formulation, network architecture and training strategies, leading to a new state-of-the-art method EQVI.

Due to the limitation of using explicit optical flow, Choi \textit{et al.} proposed CAIN \cite{cain}, which employs channel attention mechanism and PixelShuffle \cite{pixelshuffle} to directly learn the interpolated results. Recently, deformable convolutions \cite{dai2017deformable} were introduced to this task. Instead of learning pixel-wise optical flow between two frames, FeatureFlow \cite{gui2020featureflow} explored feature flows in-between corresponding deep features. Moreover, Lee \textit{et al.} \cite{adacof} proposed a new warping module named Adaptive Collaboration of Flows (AdaCoF), which can estimate both kernel weights and offset vectors for each target pixel to synthesize the output frame.

\section{Methodology}
In the following, we first revisit the quadratic video interpolation (QVI) model \cite{xu2019quadratic} in Section \ref{sec:revisit}. Then, in Section \ref{sec:RQFP}, \ref{sec:RCSN} and \ref{sec:MS}, we describe the proposed Enhanced Quadratic Video Interpolation (EQVI) method, which includes three independent components to improve the original QVI model.  Afterwards, the loss functions are introduced in Section \ref{sec:loss}. Finally, we depict the training strategies and protocols in Section \ref{sec:train}.

\subsection{Revisiting Quadratic Video Interpolation}\label{sec:revisit}
\subsubsection{Linear video interpolation.}
Given two adjacent frames $I_0$ and $I_1$, the aim of video interpolation is to generate an intermediate frame $\hat{I}_t$ at the temporal location $t$ in between the two input frames. Existing algorithms usually assume that the motion between $I_0$ and $I_1$ is uniform and linear. That is, the displacement (optical flow) of the pixel from frame 0 to $t$ can be written as a linear function:
\begin{equation}
f_{0 \rightarrow t} = t f_{0 \rightarrow 1}.
\end{equation}
However, the objects in real scenarios do not necessarily move linearly. 

\subsubsection{Quadratic video interpolation.}
\begin{figure}[t]
	\centering
	\includegraphics[width=12cm,height=4.8cm]{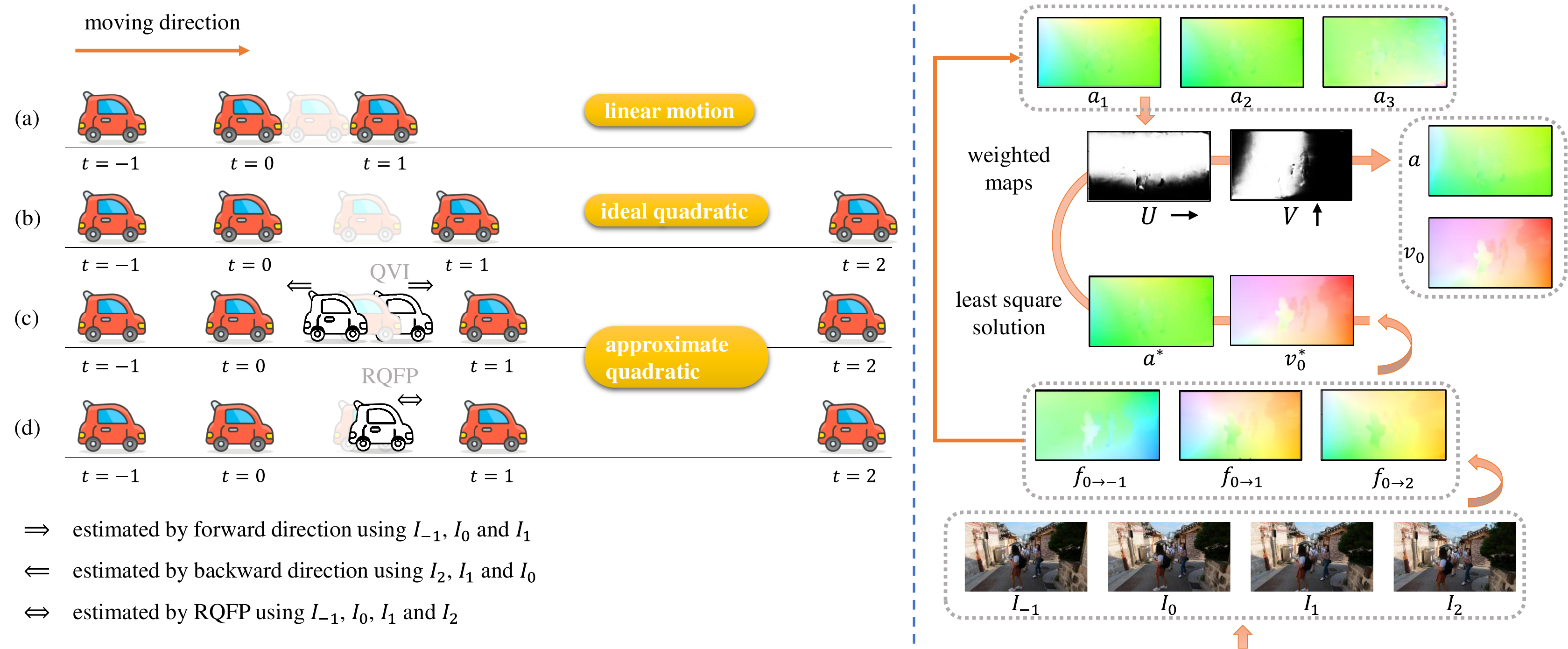}
	\caption{\textbf{Left}: Illustration for linear, quadratic and rectified quadratic flow prediction. (a) A linear motion model where an object moves at a constant velocity. (b) An ideal quadratic motion model. The motion of four input adjacent frames perfectly satisfies a uniform acceleration model. (c) When the input four consecutive frames approximately satisfy the quadratic model, the estimation of motion information in forward and backward directions would differ, due to the existence of deviations. The translucent car and the sketched car represent the ground truth intermediate frame and the estimated output frame, respectively. (d) The proposed RQFP adopts four frames to estimate the motion by least squares method (see Equation (\ref{eq:least_square})). It can obtain the solution with minimum mean square error. \textbf{Right}: An example to illustrate the working mechanism of RQFP. Given four adjacent frames, firstly, the optical flows are estimated by the flow estimation network. Then, by Equation (\ref{eq:least_square}), the least squares solutions of $a^*$ and $v_0^*$ are obtained. Meanwhile, through Equation (\ref{eq:weight}), the weighted maps $\alpha$ in $U$ (Horizontal) and $V$ (Vertical) directions are calculated. A brighter pixel in weighted map indicates the motion of this location is more likely to satisfy the quadratic model. Finally, the $a$ and $v_0$ of the quadratic model are obtained by Equation (\ref{eq:final}).		
	}
	\label{fig:illustration}
\end{figure}

Different from most previous methods which assume the motion between two input frames is linear with uniform velocity, QVI \cite{xu2019quadratic} considers higher-dimentional information, i.e. the acceleration, to describe the in-between movement (refer to Fig. \ref{fig:illustration}). Assume the acceleration is a constant, then the quadratic motion can be formulated as a uniform acceleration model:
\begin{equation}\label{eq:qua}
{f}_{0\to t} = \frac{1}{2} a t^2 + v_0 t,
\end{equation}
where $f_{0\rightarrow t}$ denotes the displacement of the pixel from frame 0 to frame $t$, $v_0$ represents the velocity at frame 0, and $a$ is the acceleration of the quadratic motion model.

\begin{figure}[h]
	\centering
	\includegraphics[width=12cm]{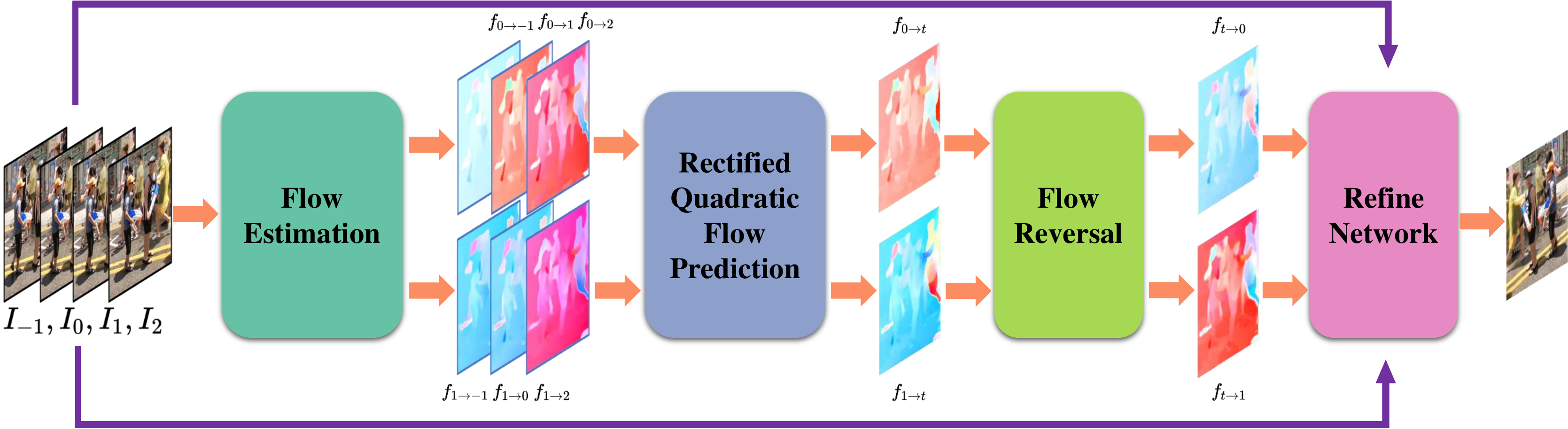}
	\caption{Quadratic video interpolation model with rectified quadratic flow prediction (RQFP). In original QVI \cite{xu2019quadratic}, the quadratic flow prediction is conducted with two flow maps ($f_{0 \rightarrow -1}$ and $f_{0 \rightarrow 1}$), while the proposed EQVI employs three flow maps ($f_{0 \rightarrow -1}$, $f_{0 \rightarrow 1}$ and $f_{0 \rightarrow 2}$) to predict the interpolated flow (forward direction).}
	\label{fig:lse-qvi}
\end{figure}

Specifically, QVI takes four consecutive frames ${I_{-1}, I_0, I_1} $ and $ {I_2}$ as inputs and predicts intermediate frame $I_t$ between $I_0$ and $I_1$ for arbitrary $t\in (0, 1)$.
The prediction process can be summarized into four steps. (The overall solution pipeline of QVI can be depicted as Fig. \ref{fig:lse-qvi}, except that the quadratic flow prediction module takes two flow maps as input).
First, four optical flows ${f}_{0\to -1}$, ${f}_{0\to 1}$, ${f}_{1\to 0}$ and  ${f}_{1\to 2}$ are estimated using PWC-Net \cite{pwc}.
Second, intermediate flows ${f}_{0\to t}$ and ${f}_{1\to t}$ are estimated by a quadratic flow prediction layer. 
Formally, if we put $t=-1$ and $t=1$ into Equation (\ref{eq:qua}), the follow equation is attained:
\begin{equation}
\label{eq:acc}
a^{qvi}={f}_{0\to 1}+{f}_{0\to -1}, \quad
v_0^{qvi}=\frac{1}{2}({f}_{0\to 1}-{f}_{0\to -1})
\end{equation}

Third, the intermediate flows are transferred to backward flows ${f}_{t\to 0}$ and ${f}_{t\to 1}$ by a flow reversal layer, and are refined by an adaptive flow filtering with a UNet-like \cite{unet} refine network. 
Finally, the intermediate $\hat I_t$ is synthesized by fusing pixels in $I_0$ and $I_1$ warped by backward flows. More details can be found in the original QVI paper \cite{xu2019quadratic}.

With the consideration of higher-dimensional information, the quadratic model can better fit the trajectories of real motions, thus predicting more accurate intermediate frames with state-of-the-art performance in benchmarks.
In this work, we use QVI as a base model and enhance it by three proposed components.

\subsection{Rectified Quadratic Flow Prediction}\label{sec:RQFP}
Based on the quadratic modelling \cite{xu2019quadratic}, we utilize a least squares method to further improve the accuracy of the quadratic flow prediction. In Equation (\ref{eq:acc}), the acceleration is estimated by three frames (or two flow maps), which could lead to the situation where the accelerations calculated in the forward and backward directions are inconsistent (see Fig. \ref{fig:illustration}(c)). Hence, we take a step further and use all four frames (or three flow maps) to estimate the quadratic model. If the motion of the input four frames basically conforms to the uniform acceleration model, the following equations should approximately hold \footnote{Put $t=-1$, $t=1$ and $t=2$ into Equation (\ref{eq:qua}), respectively.}:
\begin{equation}
\label{equ:flows}
\begin{split}
f_{0 \rightarrow -1} = -v_0 + \frac{1}{2}a, \quad
f_{0 \rightarrow 1} = v_0 + \frac{1}{2}a, \quad
f_{0 \rightarrow 2} = 2v_0 + 2a.
\end{split}
\end{equation}
The matrix form of the above equations is:
\begin{equation}
\begin{bmatrix}
-1 & 0.5 \\
1 & 0.5 \\
2 & 2 \\
\end{bmatrix}
\begin{bmatrix}
v_0  \\
a    \\
\end{bmatrix}
=
\begin{bmatrix}
f_{0 \rightarrow -1}  \\
f_{0 \rightarrow 1}    \\
f_{0 \rightarrow 2} \\
\end{bmatrix}.
\end{equation}
We can obtain the least squares solution for above overdetermined equation:
\begin{equation}\label{eq:least_square}
x^* = [A^TA]^{-1}A^Tb,
\end{equation}
where $x^*= [v_0^{*} \quad a^{*}]^T$, $A=\begin{bmatrix}
-1 & 0.5 \\
1 & 0.5 \\
2 & 2 \\
\end{bmatrix}$, $b=\begin{bmatrix}
f_{0 \rightarrow -1}  \\
f_{0 \rightarrow 1}    \\
f_{0 \rightarrow 2} \\
\end{bmatrix}$.
When the motion of input four frames approximately satisfies the quadratic assumption, the least squares solution can make a better estimation of $f_{0\rightarrow t}$ with minimum mean square error, as shown in Fig.  \ref{fig:illustration}(d). Practically we adopt several simple rules to discriminate whether the motion satisfies the quadratic assumption. Picking any two equations from Equation (\ref{equ:flows}), we can compute an acceleration $a_i$ as:
$
a_1 = f_{0 \rightarrow -1} + f_{0 \rightarrow 1} \footnote{Derived from the first and second formulas of Equation (\ref{equ:flows}). Similar derivations for the  others.},
$
$
a_2 = \frac{2}{3}f_{0 \rightarrow -1} + \frac{1}{3}f_{0 \rightarrow 2},
$
and
$
a_3 = f_{0 \rightarrow 2} + 2f_{0 \rightarrow 2}.
$
Theoretically, for quadratic motion, $a_1$, $a_2$ and $a_3$ should be in the same direction (the dot product is positive) and approximately equal to each other. If the orientations of those accelerations are not consistent, we directly utilize the original $a^{qvi}$ and $v_0^{qvi}$ in Equation (\ref{eq:acc}); otherwise, we adopt the following weighting function (adapted from $tanh$) to fuse the rectified flow and the original QVI flow:
\begin{equation}\label{eq:weight}
\alpha(z) = -\frac{1}{2}\left[ \frac{e^{\omega(z-\gamma)}-e^{-\omega(z-\gamma)}}{e^{\omega(z-\gamma)}+e^{-\omega(z-\gamma)}} \right]+\frac{1}{2},
\end{equation}
where $z=|a_1-a_2|$, $\omega$ is the axis of symmetry and $\gamma$ is the stretching factor. Empirically, we let $\omega = 5$ and $\gamma = 1$. A smaller $z$ will lead to a larger weight $\alpha(z)$, indicating that there are more motions following the quadratic formula. Then, the final estimated $v_0$ and $a$ are obtained by:
\begin{equation}\label{eq:final}
\begin{split}
v_0 = \alpha v_0^{*} + (1-\alpha) v_0^{qvi}, \quad
a = \alpha a^{*} + (1-\alpha) a^{qvi}.
\end{split}
\end{equation}
Fig. \ref{fig:lse-qvi} shows the quadratic video interpolation pipeline with rectified quadratic flow prediction module. An illustrative example of the working mechanism for RQFP is portrayed in the right part of Fig. \ref{fig:illustration}.

\begin{figure}[h]
	\centering
	\includegraphics[width=11.5cm]{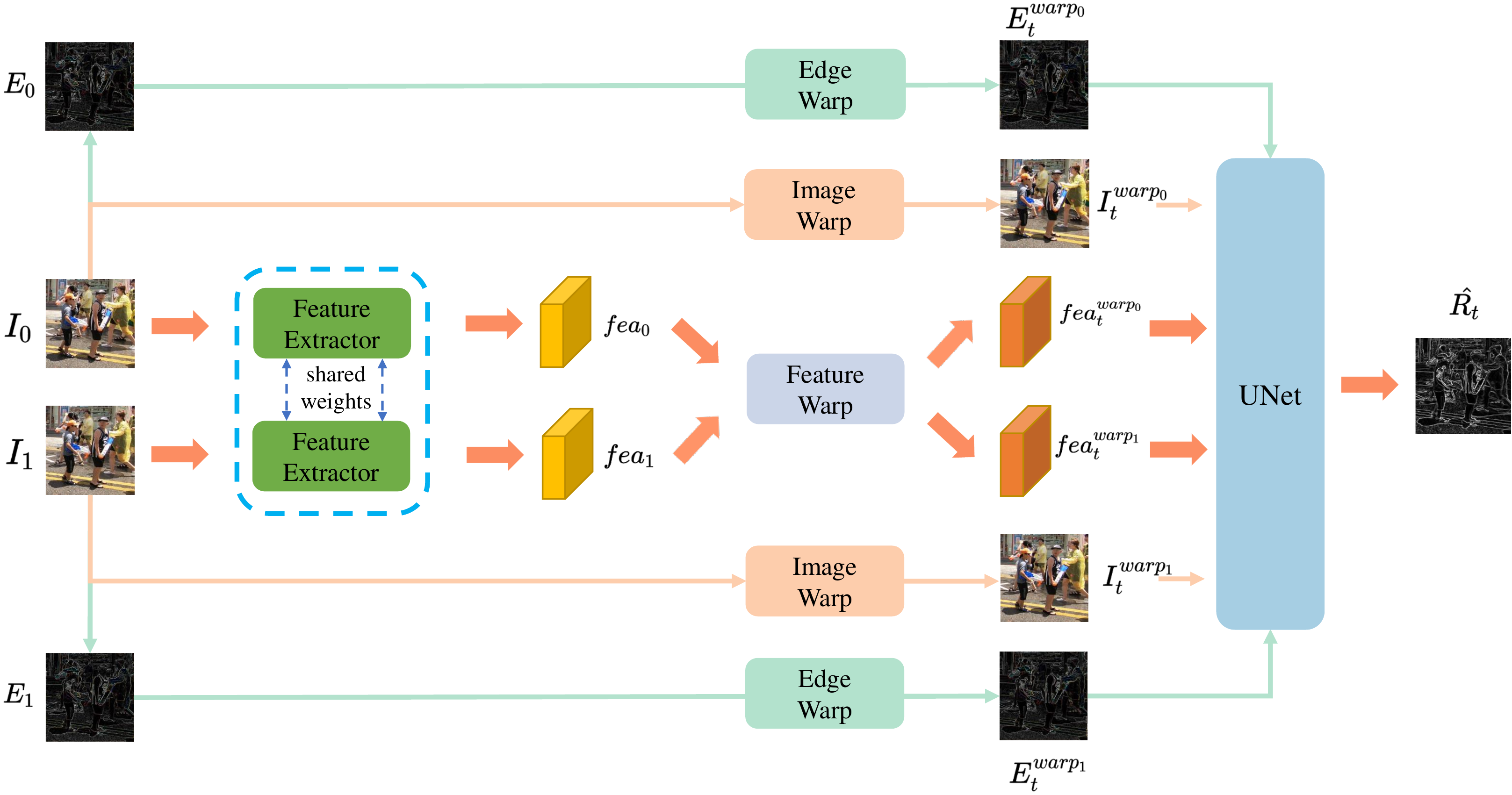}
	\caption{Residual Contextual Synthesis Network}
	\label{fig:synthesis}
\end{figure}

\subsection{Residual Contextual Synthesis Network}\label{sec:RCSN}
We introduce a residual contextual synthesis network (RCSN) to exploit the contextual information in high-dimensional feature space. The structure of RCSN is inspired by a recent video interpolation work \cite{niklaus2018context}. As shown in Fig. \ref{fig:synthesis}, we adopt the $conv1$ layer of the pre-trained ResNet18 \cite{resnet} to capture the contextual information of input frame $I_0$ and $I_1$. Then we apply backward warping on the extracted features with refined flow $f_{t \rightarrow 0}$ and $f_{t \rightarrow 1}$ to obtain pre-warped features. In addition, to preserve and leverage structural information, the edges of $I_0$ and $I_1$ are also extracted and warped. For simplicity, we calculate the gradient of each channel of the input frame as edge information. Afterwards, we feed the warped images, edges and features into a small network to synthesize a residual map $\hat{R_t}$, as depicted in Fig. \ref{fig:synthesis}. Finally, the refined output is obtained by $\hat{I}_t^{refined}=\hat{I}_t+\hat{R}_t$.

Instead of only blending pre-warped input frames, we take contextual features and edge information into account, which are complementary with the pixel-level information in image RGB space. These additional features can provide more robust and richer contextual information, which could help the model better deal with complex scenes and motion patterns. Different from \cite{niklaus2018context}, we let the context-aware synthesis network predict the residual image rather than the interpolated image. There are three main reasons for residual-learning: 1) The output of the baseline QVI has already achieved good result, so we want to predict a residual image to further refine the output, without violating the previous result. 2) Residual-learning can ease the training process and make the network converge faster. 3) By directly adding the learned residual, it is clearer to validate the performance gain obtained by the proposed module. Experiments have shown the effectiveness of the proposed component in Section \ref{sec:components}.

\begin{figure}[h]
	\centering
	\includegraphics[width=11cm]{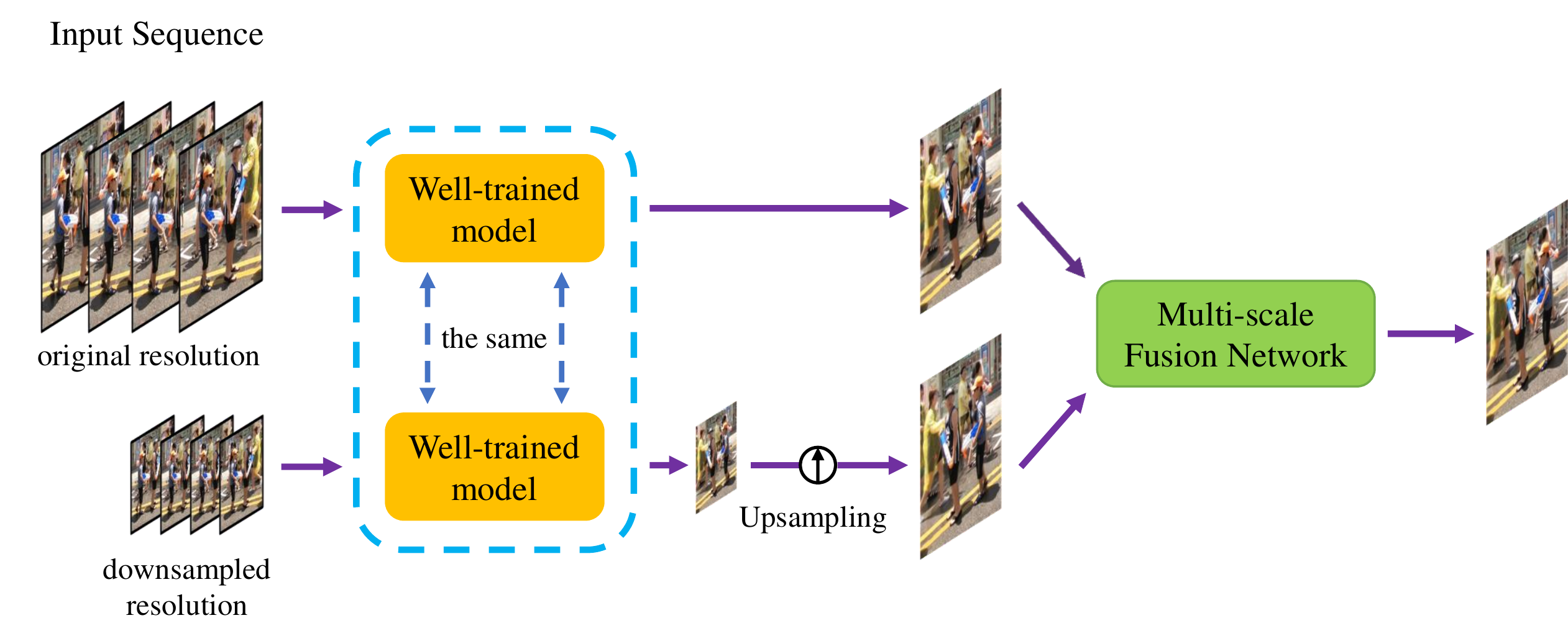}
	\caption{Multi-scale Fusion Network}
	\label{fig:ms}
\end{figure}

\subsection{Multi-scale Fusion Network}\label{sec:MS}
We propose a novel multi-scale fusion network to capture various motions at different levels. This fusion network can be regarded as a learnable augmentation process at different resolutions. As shown in Fig. \ref{fig:ms}, once we have obtained a well-trained interpolation model from previous stages, we feed the model with an input sequence $I_{in}$ and its downsampled counterpart $I_{in}^{down}$. Then a fusion network $F$ is trained to predict a pixel-wise weighted map $M$ to fuse the results.
\begin{equation}
\begin{split}
M = F(Q(I_{in}), Up(Q(I_{in}^{down}))),
\end{split}
\end{equation}
where $Q$ represents the well-trained interpolation model and $Up(.)$ denotes bilinear upsampling operations. Then, the final output frame is obtained by
\begin{equation}
\hat{I}_t^{final}=M*Q(I_{in}) + (1-M)*Up(Q(I_{in}^{down})).
\end{equation}

\subsection{Loss Functions}\label{sec:loss}
To reduce the difference between the predicted interpolated image $\hat{I}_t$ and the ground truth image $I^{gt}_t$, $L_1$ loss and the Laplacian pyramid loss $L_{lap}$ are adopted in our method. The  $L_1$ loss is as follows:
\begin{equation}
L_1 = \left\| \hat{I}_t - I^{gt}_t \right\|_1.
\end{equation}
It is a commonly-used loss function which can directly measure the distortion/fidelity between the output and the ground truth in image pixel space.

Collaborative with $L_1$ loss, we additionally adopt a Laplacian pyramid loss which is introduced by \cite{latent} for optimizing generative networks.
\begin{equation}
L_{lap} = \sum_{i=1}^{n}2^{i-1} \left\| L^i(\hat{I}_t) - L^i(I^{gt}_t) \right\|_1,
\end{equation}
where $L^i(I)$ is the $i$-th level of the Laplacian pyramid representation of an image $I$. In our implementation, the number of Laplacian pyramid layer is 5 ($n=5$). This loss is also adopted in \cite{niklaus2018context}. As a supplement to $L_1$ loss, $L_{lap}$ loss focuses more on the image edges and textures. Interestingly, experiments show that $L_{lap}$ loss has excellent potential for optimizing PSNR (more details can be found in Section \ref{sec:ab}).

\subsection{Training and Finetuning Protocols}\label{sec:train}
\begin{figure}[h]
	\centering
	\includegraphics[width=12cm]{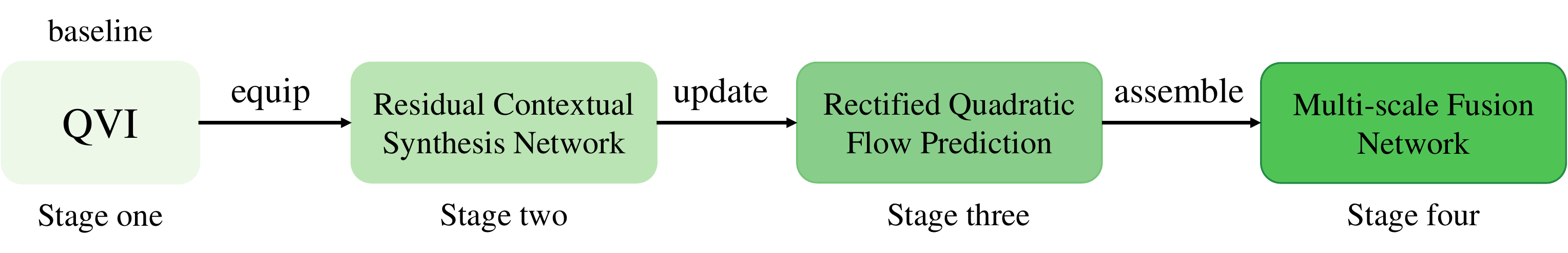}
	\caption{The overall training procedure of the proposed EQVI method}
	\label{fig:procedure}
\end{figure}

Since the proposed three components can be adopted independently, we design a four-stage training strategy during the challenge. Such stage-wise training strategy can ease the overall optimization procedure and help verify the performance gain of each component. 1) Train a baseline model QVI. We use a new flow estimation method ScopeFlow \cite{scope} instead of the original PWC-Net \cite{pwc}. The learning rate is initialized as $1\times10^{-4}$ and decayed by a factor of 10 at 100th and 150th epoch. A total of 200 epochs were executed. 2) Based on the baseline model, we add the residual contextual synthesis network (RCSN) and finetune it. The learning rate is initialized as $1\times10^{-4}$ and decayed by a factor of 10 at 15th and 25th epoch. A total of 35 epochs were executed. 3) We adopt rectified quadratic flow prediction formula (RQFP) to improve the model and finetune all the modules except optical flow estimation network. The learning rate is initialized as $1\times10^{-5}$ and decayed by a factor of 10 at 15th and 25th epoch. A total of 30 epochs were executed. 4) We input the well-trained model into the multi-scale fusion network (MS-Fusion), and finetune it. The learning rate is initialized as $1\times10^{-4}$. A total of 5 epochs were executed. Note that the rectified quadratic flow prediction has no trainable parameters. We do not adopt it in the second stage, as it requires operations on matrix and would cost more training time. The overall training procedure is depicted in Fig. \ref{fig:procedure}.

\section{Experiments}
\subsection{Training Settings and Details}
\textbf{Dataset and metrics.} Previous studies \cite{meyer2018phasenet,voxelflow,sepconv,superslomo,niklaus2018context,xu2019quadratic,adacof,gui2020featureflow} on video frame interpolation have adopted many video datasets for training and evaluating, such as Middlebury \cite{baker2011database}, UCF101 \cite{soomro2012ucf101}, Vimeo-90k \cite{xue2019video}, Adobe 240fps \cite{su2017deep} and YouTube 960-fps \cite{xu2019quadratic}. Nevertheless, every algorithm utilizes different datasets, making it difficult to conduct a fair comparison among them. To tackle this problem, REDS\_VTSR dataset is proposed for AIM 2019 VTSR Challenge \cite{Nah_2019_ICCV_Workshops_VTSR} and AIM 2020 VTSR Challenge \cite{son2020aim_VTSR} \footnote{https://data.vision.ee.ethz.ch/cvl/aim20/}. REDS\_VTSR is derived from REDS dataset \cite{Nah_2019_CVPR_Workshops_REDS}, which includes diverse scenes and motions. The input frame rate of REDS\_VTSR is 15 fps and there are two-staged target frame rates of 30 fps (interpolate one frame) and 60 fps (interpolate three frames). REDS\_VTSR consists of 240 training clips, 30 validation clips and 30 testing clips in total. We only use the 240 training clips of 60fps for training. Since the ground truth of the test set is not available, we select five representative clips from the validation set to evaluate the performance, dubbed as REDS\_VTSR5\footnote{The serial numbers are 002, 005, 010, 017 and 025.}.  For simplicity, we only evaluate the performance of interpolating one frame (the ground truth frame rate is 30 fps). PSNR and SSIM are adopted for quantitative evaluation.\\
\textbf{Implementation details.} The optical flow is estimated by ScopeFlow \cite{scope}, a recent state-of-the-art algorithm. We fix the optical flow estimation module in all training stages. The adaptive flow filtering module is kept the same as QVI \cite{xu2019quadratic}. During training, a spatial patch size of $512 \times 512$ is cropped from the original $1280 \times 720$ frames. The mini-batch size is set to 12. We use $L_1+10*L_{lap}$ loss to train the model. The training process is detailed in \ref{sec:train} and Fig. \ref{fig:procedure}. We implement our models with PyTorch framework and train them using 4 NVIDIA GeForce RTX 2080 Ti GPUs. The entire training process lasts about 6 days.

\begin{figure}[h]
	\centering
	\includegraphics[width=12.2cm]{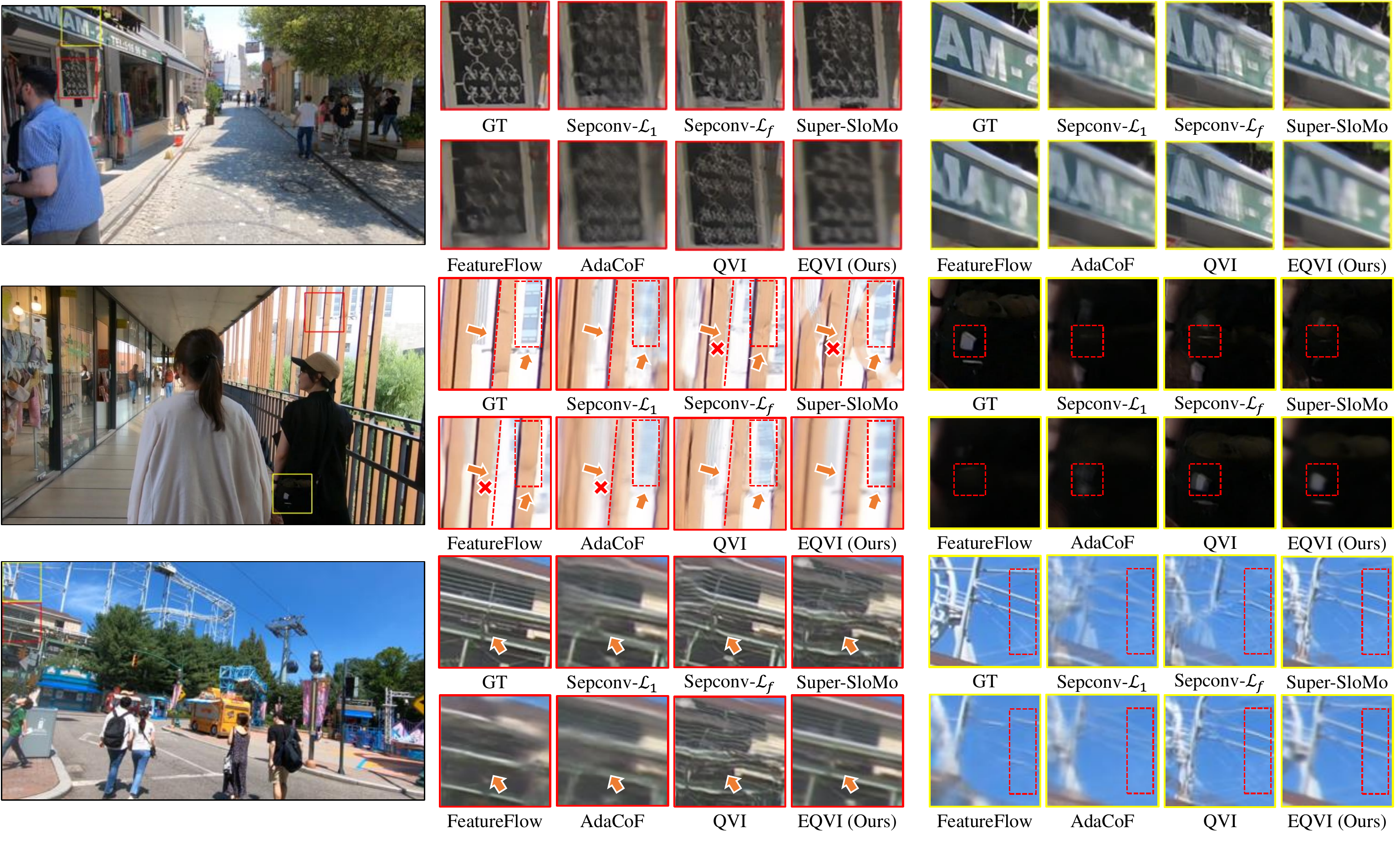}
	\caption{Qualitative comparison with other state-of-the-art methods. For images with rich textures, EQVI can generate more accurate structures and edges; for challenging images with larger motion cases, the results produced by EQVI contain fewer ghosting and artifacts. Please see the highlighted marks in the figure.}
	\label{fig:compare}
\end{figure}

\subsection{Effectiveness of the Proposed Components} \label{sec:components}
To facilitate the training process and demonstrate the effectiveness of each component, we introduce a four-stage training strategy. The evaluation results of each stage are listed in Table \ref{Tab:components}. The baseline model achieves 24.75dB on the REDS\_VTSR5 dataset. After equipping the residual contextual synthesis network, the performance reaches 24.93dB, achieving an improvement of 0.18dB, which reveals the great potential of exploiting the high-dimensional contextual information in feature space. Then, we update the model with the rectified quadratic flow prediction as described in Section \ref{sec:RQFP}, and the performance can be further boosted to 24.96dB. In the fourth stage, the previously trained model is put into the multi-scale fusion network, after which the results can reach 24.97dB. However, we find that the utilization of the multi-scale fusion network could decrease the performance of SSIM. In summary, the experimental results have shown that the proposed components have greatly improved the performance of the original QVI.

After the competition, we also train a model from scratch with RCSN and RQFP equipped. Interestingly, the results show that training from scratch can yield a better performance with 25.069dB on REDS\_VTSR5 dataset. However, it costs more training time (about 9 days). From the experiments, it can be concluded that the proposed method is very effective, flexible and easy to train.

\setlength{\tabcolsep}{4pt}
\begin{table}[h]
	\begin{center}
		\caption{Test results on the REDS\_VTSR5 dataset. Based on the standard QVI baseline, the proposed module improves the performance step by step.}
		\label{Tab:components}
		\begin{tabular}{lcccc|cc}
			\hline
			& baseline & RCSN & RQFP & MS-Fusion & PSNR & SSIM\\
			\hline
			Stage one  & $\checkmark$ &  & & & 24.746 & 0.717\\
			Stage two  & $\checkmark$ & $\checkmark$ & & & 24.927 & 0.725\\
			Stage three  & $\checkmark$ & $\checkmark$ & $\checkmark$ & & 24.963 & 0.727\\
			Stage four  & $\checkmark$ & $\checkmark$ & $\checkmark$ & $\checkmark$ & 24.971 & 0.726\\
			\hline
			From scratch  & $\checkmark$ & $\checkmark$ & $\checkmark$ & $\times$ & 25.069 & 0.729\\
			\hline
		\end{tabular}
	\end{center}
\end{table}
\setlength{\tabcolsep}{1.5pt}

\subsection{Comparison with the State-of-the-arts}
We compare our EQVI model with several state-of-the-art methods on video interpolation: Sepconv \cite{sepconv}, Super-SloMo \cite{superslomo}, QVI \cite{xu2019quadratic}, AdaCoF \cite{adacof} and FeatureFlow \cite{gui2020featureflow}. However, most of these methods utilize different datasets for training and testing, making the comparison difficult. To be fair, we retrain QVI and AdaCoF with REDS\_VTSR training set. For Sepconv, Super-SloMo and FeatureFlow, we just use their released pretrained models. 

Quantitative results are shown in Table \ref{Tab:comparison}. Our EQVI outperforms all the state-of-the-art methods by a large margin. Specifically, compared with the standard QVI model which achieves the second-best result, we improve the performance by about 0.38dB on the REDS\_VTSR5 dataset. The visualization comparisons in Fig. \ref{fig:compare} and Fig. \ref{fig:pageone} show that EQVI has its unique merits of dealing with large and complex motion. It can produce more accurate interpolated results with higher PSNR values while preserving the image textures and structures. To be specific, Sepconv, Super-SloMo, FeatureFlow and AdaCoF cannot model and estimate complex frame motions well, leading to inaccurate and blurry interpolated results with severe ghosting and artifacts. Compared with standard QVI, after equipping with three aforementioned components, the predicted images are further enhanced with more accurate textures and fewer ghosting artifacts.

\begin{minipage}[h]{1\textwidth}
	\begin{minipage}[t]{0.45\textwidth}
		\centering
		\makeatletter\def\@captype{table}\makeatother\caption{Quantitative comparison on REDS\_VTSR5. EQVI achieves the best performance.}
		\setlength{\tabcolsep}{3pt}
		\label{Tab:comparison}
		\begin{tabular}{lll}
			\hline\noalign{\smallskip}
			Method & PSNR & SSIM\\
			\hline
			Sepconv-$\mathcal{L}_{1}$  \cite{sepconv}& 23.123 & 0.662\\
			Sepconv-$\mathcal{L}_{F}$ \cite{sepconv}& 22.770 & 0.641\\
			Super-SloMo \cite{superslomo} &  23.004 & 0.659\\
			FeatureFlow \cite{gui2020featureflow} & 22.990 & 0.666
			\\
			AdaCoF \cite{adacof} & 23.462 & 0.675\\
			QVI \cite{xu2019quadratic} & \textcolor[RGB]{0,0,255}{{24.591}}  & \textcolor[RGB]{0,0,255}{{0.710}}\\
			EQVI & \textcolor[RGB]{255,0,0}{\textbf{24.971}} & \textcolor[RGB]{255,0,0}{\textbf{0.726}}\\
			\hline
		\end{tabular}
	\end{minipage}
	\begin{minipage}[t]{0.45\textwidth}
		\centering
		\makeatletter\def\@captype{table}\makeatother\caption{Top5 methods in the AIM2020 VTSR Challenge. Our method won the first place.}
		\label{Tab:AIM2020}
		\setlength{\tabcolsep}{2.6pt}
		\begin{tabular}{c|cc|cc}
			\hline
			\multirow{2}{*}{Method} & \multicolumn{2}{c}{15fps $\rightarrow$ 30fps } &    \multicolumn{2}{|c}{15fps $\rightarrow$ 60fps} \\
			& \small\underline{PSNR} &  \small\underline{SSIM} & \small\underline{PSNR} & \small\underline{SSIM} \\
			\hline
			Ours & \textcolor{red}{\textbf{24.78}} & \textcolor{blue}{{0.712}} & \textcolor{red}{\textbf{25.69}} & \textcolor{blue}{{0.742}}\\
			2nd & \textcolor{blue}{{24.69}} & \textcolor{red}{\textbf{0.714}} & \textcolor{blue}{{25.61}} & \textcolor{red}{\textbf{0.746}}\\
			3rd & 24.40 & 0.697 & 25.19 & 0.727\\
			4th & 24.29 & 0.698 & 25.05 & 0.727\\
			5th & 23.59 & 0.672 & 24.36 & 0.700\\
			\hline
		\end{tabular}
	\end{minipage}
\end{minipage}

\subsection{Results of AIM2020 VTSR Challenge}
We participated in the AIM2020 VTSR Challenge \cite{son2020aim_VTSR} with the proposed EQVI model and won the first place. The test results are ranked by PSNR in Table \ref{Tab:AIM2020}. In both $15fps \rightarrow 30fps$ (interpolate one frame) and $15fps \rightarrow 60fps$ (interpolate three frames) tracks, our EQVI achieves the highest PSNR values.

\subsection{Ablation Study} \label{sec:ab}
\subsubsection{Loss combination.} In the main experiments, we combine $L_1$ and $L_{lap}$ loss functions together and achieve appealing performance in terms of PSNR and SSIM. To validate the effectiveness of different loss combinations, we conduct an ablation study using various losses to train the baseline model: 1) Only adopt $L_1$ loss. 2) Only adopt $L_{lap}$ loss. 3) Combine $L_1$ and $L_{lap}$ losses together. As shown in Table \ref{Tab:loss}, adopting Laplacian loss $L_{lap}$ can obtain better PSNR values than $L_1$ loss, but would decrease the performance of SSIM. Moreover, combining $L_1$ and $L_{lap}$ can achieve higher PSNR, and to some extent, make up for the decrease of SSIM caused by adopting $L_{lap}$ loss.

\subsubsection{Effectiveness of RQFP.}To directly demonstrate the effectiveness of the proposed rectified quadratic flow prediction formula, comparison experiments are conducted based on the well-trained model in stage two (see Fig. \ref{fig:procedure}): 1) Finetune the model without RQFP. 2) Finetune the model with RQFP. The results are depicted in Table \ref{Tab:RQFP}. The initial performance of the model in stage two is 24.93dB. After equipping the proposed RQFP, the performance improves to 24.96dB while the performance keeps nearly the same without RQFP.

\begin{minipage}[t]{\textwidth}
	\begin{minipage}[t]{0.45\textwidth}
		\centering
		\makeatletter\def\@captype{table}\makeatother\caption{Ablation study on the loss combination.}
		\label{Tab:loss}
		\setlength{\tabcolsep}{8pt}
		\begin{tabular}{ccc}
			\hline\noalign{\smallskip}
			loss & PSNR &  SSIM \\
			\noalign{\smallskip}
			\hline
			\noalign{\smallskip}
			$L_1$ & 24.646 & \textbf{0.718}\\
			$L_{lap}$ & 24.746 & 0.716\\
			$L_{lap} + L_1$ & \textbf{24.775} & 0.717\\
			\hline
		\end{tabular}
	\end{minipage}
	\begin{minipage}[t]{0.48\textwidth}
		\centering
		\makeatletter\def\@captype{table}\makeatother\caption{By finetuning with RQFP, the result is further boosted.}
		\label{Tab:RQFP}
		\setlength{\tabcolsep}{3pt}
		\begin{tabular}{ccc}
			\hline\noalign{\smallskip}
			model & PSNR &  SSIM \\
			\noalign{\smallskip}
			\hline
			\noalign{\smallskip}
			stage-two model & 24.927 & 0.725\\
			finetune w/o RQFP & 24.933 & 0.725\\
			finetune w/ RQFP & \textbf{24.963} & \textbf{0.727}\\
			\hline
		\end{tabular}
	\end{minipage}
\end{minipage}

\subsubsection{Exploration on RCSN.} In Section \ref{sec:RCSN}, we discussed the proposed residual contextual synthesis network (RCSN), which integrates pre-warped features, images and edges together to exploit the contextual information as shown in Fig. \ref{fig:synthesis}. In this session, we further validate the effectiveness of such contextual information. As shown in Table \ref{Tab:RCSN}, if we only feed pre-extracted features in RCSN, the PSNR and SSIM are 24.916dB and 0.724, respectively. After adding pre-warped images into RCSN, the performance improves to 24.925dB and 0.725. When edge information is further employed, the PSNR reaches 24.927dB.

\begin{minipage}[t]{\textwidth}
	\begin{minipage}[t]{0.45\textwidth}
		\centering
		\makeatletter\def\@captype{table}\makeatother\caption{Exploration on RCSN.}
		\label{Tab:RCSN}
		\setlength{\tabcolsep}{2.4pt}
		\begin{tabular}{ccc}
			\hline\noalign{\smallskip}
			contextual info. & PSNR &  SSIM \\
			\noalign{\smallskip}
			\hline
			\noalign{\smallskip}
			features & 24.916 & 0.724\\
			features+images & 24.925 & 0.725\\
			features+images+edges & \textbf{24.927} & \textbf{0.725}\\
			\hline
		\end{tabular}
	\end{minipage}
	\begin{minipage}[t]{0.5\textwidth}
		\centering
		\makeatletter\def\@captype{table}\makeatother\caption{Influence of optical flow.}
		\label{Tab:flow}
		\setlength{\tabcolsep}{2pt}
		\begin{tabular}{ccc}
			\hline\noalign{\smallskip}
			flow model & PSNR &  SSIM \\
			\noalign{\smallskip}
			\hline
			\noalign{\smallskip}
			PWC-Net \cite{pwc}& 24.591 & 0.710\\
			ScopeFlow \cite{scope}& \textbf{24.746} & \textbf{0.716}\\
			\hline
		\end{tabular}
	\end{minipage}
\end{minipage}

\subsubsection{Influence of optical flow estimation.} In flow-based methods, the accuracy of flow estimation is crucial. To validate the effects of different flow estimation algorithms in our method, we train the baseline model with two state-of-the-art algorithms – PWC-Net \cite{pwc} and ScopeFlow \cite{scope}. In the optical flow benchmarks, ScopeFlow is superior to PWC-Net. As expected, as shown in Table \ref{Tab:flow}, the baseline model equipped with ScopeFlow also outperforms that with PWC-Net, which shows the paramount importance of optical flow estimation algorithm in video frame interpolation task.

\section{Conclusions}
We have presented the winner solution of AIM2020 Video Temporal Super-Resolution Challenge. We propose an enhanced quadratic video interpolation (EQVI) model from the aspects of formulation, network architecture and training strategies. We adopt least squares method to rectify the estimation of motion flow. A residual contextual synthesis network is introduced to employ contextual information in high-dimensional feature space. In addition, we devise a novel learnable augmentation process -- multi-scale fusion network to further boost the performance. The proposed EQVI model outperforms all recent state-of-the-art video interpolation methods quantitatively and qualitatively.

\section{Acknowledgement}
This work is partially supported by the National Natural Science Foundation of China (61906184), Science and Technology Service Network Initiative of Chinese Academy of Sciences (KFJ-STS-QYZX-092), Shenzhen Basic Research Program (JSGG20180507182100698, CXB201104220032A), the Joint Lab of CAS-HK，Shenzhen Institute of Artificial Intelligence and Robotics for Society.

%
%
\bibliographystyle{splncs04}
\bibliography{egbib}
\end{document}